\documentclass[pdflatex,sn-mathphys-num]{sn-jnl}

\usepackage{graphicx}%
\usepackage{multirow}%
\usepackage[table,xcdraw]{xcolor}
\usepackage{float}
\usepackage{lineno,hyperref,subfig,multirow,booktabs}
\usepackage[noend]{algpseudocode}
\usepackage{amsmath,amssymb,amsfonts}%
\usepackage{amsthm}%
\usepackage{mathrsfs}%
\usepackage[title]{appendix}%
\usepackage{xcolor}%
\usepackage{textcomp}%
\usepackage{manyfoot}%
\usepackage{booktabs}%
\usepackage{algorithm}%
\usepackage{algorithmicx}%
\usepackage{algpseudocode}%
\usepackage{listings}%

\theoremstyle{thmstyleone}%
\theoremstyle{thmstyletwo}%

\theoremstyle{thmstylethree}%

\begin{document}

\title[Article Title]{Future Trends in the Design of Memetic Algorithms: the Case of the Linear Ordering Problem}


\author*[1]{\fnm{Lázaro} \sur{Lugo}}\email{lazaro.perez@cimat.mx}
\author[1]{\fnm{Carlos} \sur{Segura}}\email{carlos.segura@cimat.mx}
\author[2]{\fnm{Gara} \sur{Miranda}}\email{gmiranda@ull.edu.es}

\equalcont{These authors contributed equally to this work.}

\affil[1]{\orgdiv{Área de Computación}, \orgname{Centro de Investigación en Matemáticas (CIMAT)}, \orgaddress{\street{Callej\'on Jalisco s/n}, \city{Guanajuato}, \postcode{36240}, \state{Guanajuato}, \country{México}}}
\affil[2]{\orgdiv{Departamento de Ingeniería Informática y de Sistemas}, \orgname{Universidad de La Laguna}, \orgaddress{\street{Camino San Francisco de Paula 19}, \city{Santa Cruz de Tenerife}, \postcode{38200}, \state{Canarias}, \country{España}}}


\abstract{
The way heuristic optimizers are designed has evolved over the decades, as computing power has increased.
Such has been the case for the Linear Ordering Problem (LOP), a field in which trajectory-based strategies led the way during the 1990s, but which have now been surpassed by memetic schemes.
This paper focuses on understanding how the design of LOP optimizers will change in the future, as computing power continues to increase, yielding two main contributions.
On the one hand, a metaheuristic was designed that is capable of effectively exploiting a large amount of computational resources, specifically, computing power equivalent to what a recent core can output during runs lasting over four months.
Our analyses show that as the power of the computational resources increases, it will be necessary to boost the capacities of the intensification methods applied in the memetic algorithms to keep the population from stagnating.
And on the other, the best-known results for today's most challenging set of instances (xLOLIB2) were significantly outperformed. New bounds were established in this benchmark, which provides a new frame of reference for future research.
}

\keywords{Linear Ordering Problem, Memetic Algorithms, High-performance Capabilities, Design of Optimizers}



\maketitle

\section{Introduction}
\label{sec:introduction}

An effective decision-making process usually involves solving highly complex optimization problems.
%
The most general classification of optimization techniques distinguishes between exact and approximate optimization~\cite{blum2003metaheuristics}. 
Given the complexity of many real-life optimization problems, exact optimization methods are usually not applicable within reasonable timeframes.
Therefore, approximate methods have gained considerable popularity~\cite{li2022heuristics}. 
In this area, metaheuristics offer the possibility of increasing the generality of the developments, thus reducing the design efforts.

One highly complex problem that has benefited greatly from the use of metaheuristics is the Linear Ordering Problem (LOP), which is an NP-hard combinatorial optimization problem with a large number of applications.
The simplest definition of it is based on the matrix triangulation problem. 
Given a matrix $M_{n \times n} = (m_{ij})$, the problem is to determine a simultaneous permutation $\sigma$ of the rows and columns of $M$ such that the sum of the entries above the main diagonal is maximized~\cite{Marti:2011}. 
As a result, the problem can be formulated as the search for a permutation $\sigma$ that maximizes (\ref{eq:lop}): 
 
\begin{equation}
\label{eq:lop}
 \sum_{i = 1}^{n - 1} \sum_{j = i + 1}^{n} m_{\sigma_{i}\sigma_{j}}
\end{equation}

LOP has been applied in numerous areas, such as: determination of ancestry relationships in archaeology \cite{Glover:1974}, straightline crossing minimization~\cite{Junger:1997}, 
and
the development of probabilistic decision-making models \cite{SMEULDERS201832}. 
Moreover, since many of these applications involve the use of large matrices, highly complex test sets have been defined.
In particular, the most challenging validation set today is the LOLIB library~\cite{Marti:2012}, the latest version of which includes instances up to size 1000.

Many diverse algorithms have been proposed to deal with the LOP~\cite{Marti:2011}.
Since there are no specific guides of steps to follow in the design of optimizers, researchers draw on the practical experience gained from similar problems~\cite{Talbi:09}.
In addition, the relative efficiency between metaheuristics depends largely on the computational capabilities available and the complexity of the instance to be solved, which accounts for the need for a wide variety of proposals~\cite{michel2005}.
%
%
The LOP is a fairly clear example of how computational resources and the complexity of instances affect the design of optimizers.
Based on the historical review presented in~\cite{Marti:2011}, it is clear that effective optimizers have been evolving from ad-hoc heuristic methods, to trajectory metaheuristics, and on to the current population-based metaheuristics~\cite{Ceberio:2015, lugo2022diversity}.

This paper falls within the field of approximate optimization and aims to contribute to the advancement of the design and development of efficient and robust optimizers. 
It focuses on the case of the LOP and seeks to answer the following research questions:

\begin{enumerate}

    \item Are the methods that generate the best solutions for the LOP able to do so in reasonable times, for example, under a day in the most challenging instances?
    
    \item Will today's cutting-edge methods in the LOP continue to retain their prominence in the computing systems of the future, that is, after a significant increase in computing capabilities? Or, on the contrary, will they be unable to scale properly?
       
    \item Have optimal solutions already been achieved for the current most challenging datasets, or, on the contrary, is there still room for improvement?
   
\end{enumerate}

In order to answer these questions at this time, it is necessary to make use of long-term executions and parallelizations to simulate the increased availability of computational resources.
With this goal in mind, we conducted a long-term performance analysis of the current leading method for the LOP, the MA-EDM algorithm~\cite{lugo2022diversity}.
This analysis allowed us to identify some shortcomings in this method, and led us to conclude that, once the computing power is increased, MA-EDM has to be redesigned in order to leverage the additional computing power.
In light of the weaknesses identified in the MA-EDM, a new optimizer called Memetic Algorithm with Explicit Diversity Management and Enhanced Intensification (MA-EDM$ei$) is proposed that applies an Iterated Local Search (ILS) as an improvement mechanism.
As demonstrated in the experimental study, in high-performance environments, MA-EDM$ei$ vastly outperforms the current best strategy.
Moreover, by parallelizing MA-EDM$ei$, we were able to simulate scenarios with computing capabilities equivalent to those of a state-of-the-art core in executions lasting over 4 months. 
This experiment revealed that the best solutions found so far exhibit a significant gap with respect to the optimal ones, indicating that this set of instances has not yet been optimally solved, meaning, therefore, that more methodological advances are required. 
As an additional contribution, it should be noted that the new Best-Known Solutions (BKS) obtained for instances ranging in size from 300 to 1,000 will provide a frame of reference that will allow other researchers to better evaluate future advances.

The rest of the document is organized as follows. 
Section \ref{sec:futureLimitationsMA-LOP} discusses the design of approximate algorithms, with a special emphasis on Memetic Algorithms and their adaptations to the LOP.
Section \ref{sec:usingHighPerformanceCE} focuses on describing the use of high-performance computing environments to achieve the objectives of this research.
Section \ref{sec:adaptationofMAEDM} justifies and presents MA-EDM$ei$.
The experimental validation is presented in Section \ref{sec:ExperimentalValidation}.
Finally, the last section, discusses the conclusions and future areas of work.


\section{Future and Limitations of Memetic Algorithms for the LOP}
\label{sec:futureLimitationsMA-LOP}

Evolutionary Algorithms (EAs) are a class of metaheuristics that draw their inspiration from biological evolution~\cite{Cotta:2018}.
EAs are very effective at solving difficult optimization problems in a wide variety of fields~\cite{dasgupta2013evolutionary}.
However, they typically become more effective when combined with refinement mechanisms~\cite{Neri:2011:handbook}, combination known as Memetic Algorithms (MAs)~\cite{NERI20121}.
In first-generation MAs, enhancement mechanisms were used statically to refine the solutions.
Subsequently, a variety of more complex integration strategies were developed, ranging from schemes that consider a set of improvement mechanisms that are applied adaptively, to schemes that rely on the feedback provided by the improvement mechanisms to adapt internal parts of the EA.
Despite the advances offered by recent MAs~\cite{Zheng:23}, their design, use and implementation is very complex due to the number of parameters and components they consider.
In many problems, the first-generation MAs are still effective~\cite{fisher2007, Ceberio:2015}, so both the simplest and the most complex methodologies continue to be investigated.
Considering the number of problems in which MAs are the leading strategy~\cite{Cotta:2018}, and that this is the case with the LOP, this paper focuses on these types of strategies.
In particular, the MA that has yielded the current BKS for the LOP is a first-generation algorithm~\cite{lugo2022diversity}, so this is the main method used to conduct our analyses.

%
%
%
Since, for many problems, finding a local optimum is a computationally very expensive process, in such cases it is not feasible to use more complex schemes, which is why local search remains one of the most used methods in MAs; however, advances in hardware have also made it quite common for more complex trajectory metaheuristics and even mathematical programming strategies to be considered. Thus, there is a tendency to consider more complex refinement strategies, as the computation capabilities increase.
In relation to the design of MAs, the problem of how often and how the improvement mechanism is used is a fundamental task that has been addressed in the literature in various ways~\cite{oliveto2007time}. 
Another crucial point is the definition of the neighborhood for the local search.
In order to aid in the selection of the neighborhood, some heuristic procedures for analyzing the fitness landscape have been devised~\cite{NERI20121}.
%
It is also important to note that in addition to the design issues implicit to MAs, as with any population-based algorithm, there are several additional characteristics that play an important role in the success of the optimization process.
In particular, proper diversity management plays a critical role~\cite{Crepinsek:13}.
%
%
%
 
In the case of MAs, problems with diversity management appear frequently.
For example, in the RHGA algorithm~\cite{Zheng:23}, one of the most effective strategies for the TSP, great care had to be taken in selecting the frequency of application of the LKH improvement strategy to avoid premature convergence.
This type of problem arises especially in the case of high dimensionality problems~\cite{Segura:15}, so as the instances grow, it is quite common to have to change some design decisions.
Issues related to instances of high dimensionality are not unique to MAs~\cite{chen2015measuring}.
This problem is known as the ``curse of dimensionality'',
%
which occurs when an increase in search space size causes a significant loss of performance in optimization algorithms.
The increase in the dimensionality of the instance causes the search space to grow exponentially in most problems, and as a result, the percentage of solutions evaluated decreases.
However, this is not its only drawback, since it can, for example, cause premature convergence in subsets of variables~\cite{Segura:15}.
Therefore, increasing the amount of computing power available (even exponentially) may not solve the challenges that arise when dealing with large instances.
To address this problem, various techniques, such as dimensionality reduction techniques and coevolution, have been proposed~\cite{Wang:2021}.

This paper takes advantage of the demands associated with high dimensionality problems to further analyze and extend the metaheuristic schemes known to perform well with the LOP.
To do this, we will work with instances of up to $1,000$, that is, with a search space with approximately $10^{2567}$ candidate solutions, and we will give the algorithm what it supposedly needs: more computing power. 
It is not at all foolhardy to think that in the near future, computing capacity will advance significantly, which will make more computing power available to algorithms. 
However, what we do not know is whether today’s methods will be able to continue to perform well when these resources are made available to them.
This research provides an answer to this question.


\section{Using High Performance Computing Environments}
\label{sec:usingHighPerformanceCE}

High performance computing (HPC) laboratories are typically used to reduce the time required to solve highly complex problems.
Since the programming and bottlenecks of parallel algorithms differ from those of sequential algorithms, it is quite common for parallel algorithms to be developed that are significantly different from the direct parallelizations of the best sequential algorithms.
Thus, in the case of Memetic Algorithms and population-based optimizers in general, various techniques have emerged to take advantage of this type of resource~\cite{Albaparallel2012}, such as master-worker parallelizations, island models, coevolution and cellular models.
Among them, the only strategy that maintains the panmictic population-based model of the sequential algorithm is the master-worker model.

The objective of using HPC in this research was not to take advantage of parallel computing resources to obtain high-quality results quickly, but to understand how the effectiveness of one of the leading algorithms in the area of LOP is affected by the increase in computing capabilities, as well as to propose modifications that allow the increase in computing power to be better utilized.
Thus, the purpose of this work is not to innovate in parallelization models or find the best parallel models for the LOP, but to generate new knowledge in relation to the optimization of the LOP.
In order to properly attain this objective, models should be used that keep the evolutionary process intact.
Because of this, the synchronous master-worker model was selected.
This scheme is standard in the parallelization of population-based algorithms and it based on the redistribution of a set of independent tasks among a set of processes that are exclusively dedicated to executing those tasks.

\begin{algorithm}[!t]
\caption{Memetic Algorithm with Explicit Diversity Management (MA-EDM)}
\label{alg:MA_EDM}
\begin{algorithmic}[1]
\Require {\textit{LOP instance}, population size: \textit{N}, \textit{stopping criterion: $time$}}
\State{\textbf{Initialization:} Generate an initial population $P_0$ with $N$ individuals, assign $i=0$}
\State{\textbf{Intensification:} Execute local search in $P_0$}
\State{\textbf{Diversity Initialization:} Calculate the minimum desired initial distance ($D_0$) as the average distance 
between individuals in $P_0$}
\While {stopping criterion ($time$) not reached}
	\State{\textbf{Parent selection:} apply binary tournament in $P_i$ to fill the parent set with $N$ parents}
	\State{\textbf{Crossover:} apply $CX$ in the parent set to obtain the set $O_i$ with $N$ offspring}
	\State{\textbf{Intensification:} execute local search in $O_i$}
	\State{\textbf{Replacement:} apply $BNP(P_i,O_i)$, which takes into account the diversity, to create $P_{i+1}$}
	\State{$i=i+1$}\EndWhile
\State{\textbf{Return} best evaluated solution}
\end{algorithmic}
\end{algorithm}

The MA-EDM sequential algorithm is a standard first-generation MA.
Its general behavior is described in Algorithm~\ref{alg:MA_EDM}.
A detailed description of each of its components is available at~\cite{lugo2022diversity}.
Each iteration operates with a population of $N$ individuals and consists of a binary tournament-based parent selection phase, a child-reproduction phase that considers the Cycle Crossover (CX), an improvement phase that applies a local search, and a replacement phase that uses the Best-Non Penalized strategy (BNP).
Of all the above steps, the local search is the most computationally demanding process, since the neighborhood is considered based on shifts~\cite{Musliu:2004}, so for instances of size ${n}\times{n}$, the step of ensuring that a local optimum has been found using incremental evaluation has a computational cost of $O(n^2)$ while the rest of the components depend linearly or sublinearly on $n$.
Therefore, with large instances like the ones considered in this research, it is natural for the improvement process to be the task that is distributed in the parallelization.
This distribution is done synchronously so as not to change the evolutionary process of the sequential scheme; that is, once a local search distribution point (or another improvement procedure) is reached, the master process halts until all the local searches are complete.

\begin{figure*}[t] 
\centering 
\includegraphics[width=0.9\textwidth]
{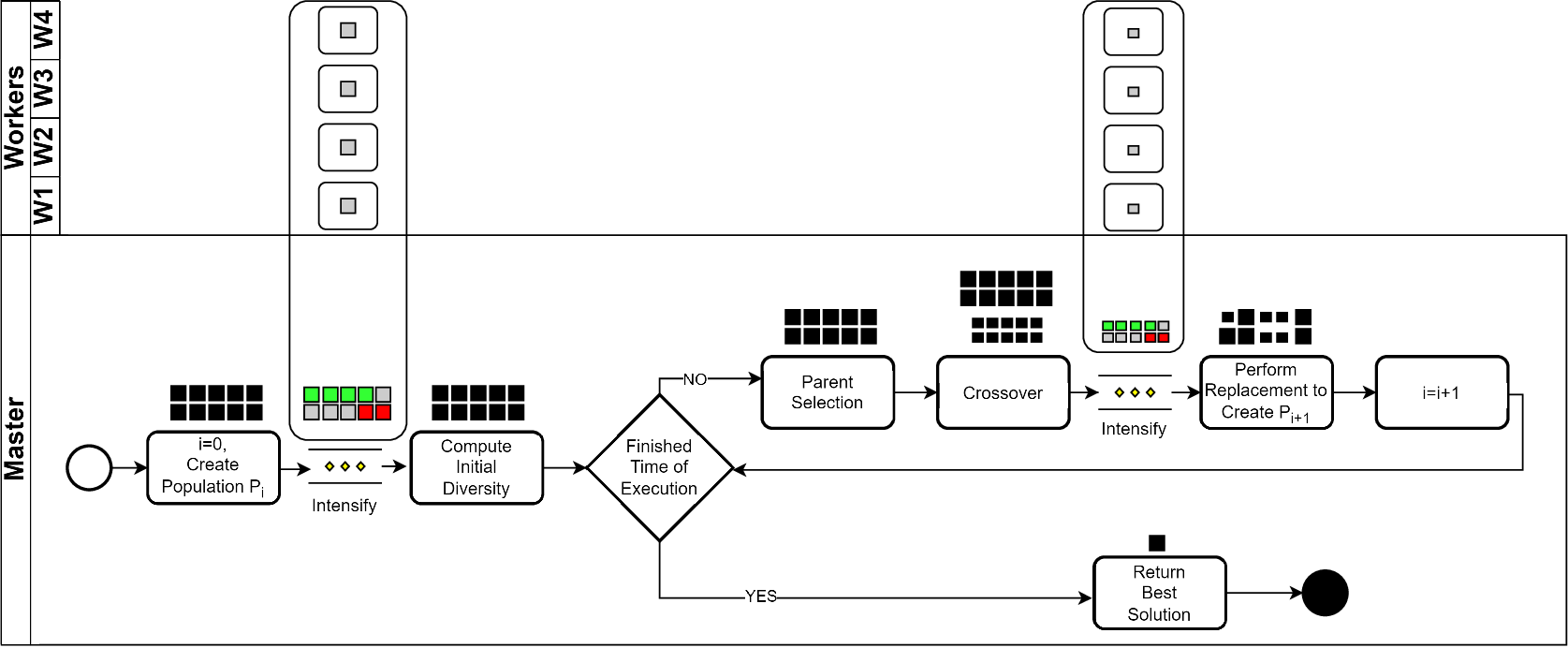}
\caption{Flowchart of the \textit{PMA-EDM} algorithm} 
\label{fig:flow_chart_MAEDMILSP}
\end{figure*}

The flowchart shown in Figure~\ref{fig:flow_chart_MAEDMILSP} illustrates the operation of the parallel version of MA-EDM (PMA-EDM).
The diagram shows an example with a population of size 10, in which one master process (bottom) and four worker processes (top) are executed.
The population is represented with larger squares, while the offspring is represented with smaller squares.
The rectangular blocks in the region of the master process are executed sequentially, while the procedures performed in parallel are represented using two horizontal lines, and extend to the region corresponding to the workers.
There are two stages in which the intensification procedure is applied to a set of individuals: after creating the initial population, and after creating the offspring.
Those two blocks are the ones that are executed in parallel.
By using the Message Passing Interface (MPI), the individuals are distributed so that the worker processes carry out the intensification process and output the results.
This distribution is carried out dynamically, meaning we wait for a process to finish carrying out the current intensification before sending it a new individual, continuing with this strategy as long as there are individuals remaining to be intensified.
In the parallelized blocks, i.e., in the ``Intensify'', the individuals that have already been sent to the work processes and were intensified are shown in green, those in which the intensification process is taking place are gray, and those that have not yet been intensified are red.
The process will continue until the intensification has been performed on all the individuals, at which point the master process will continue with the next sequential blocks.
The advantage of doing it dynamically is that it avoids imbalance problems when the tasks are heterogeneous. 
Such is the case with the local search.
Given that in the case of the LOP, the degree of heterogeneity is not high, in this research we selected population sizes that are multiples of the number of workers, which is an important consideration to enhance the efficiency.
Finally, it is important to note that this way of parallelizing is independent of the components selected to carry out the intensification process, which is why the flowchart is presented in a general way. 
The parallelization shown was used to parallelize an extension of the MA-EDM algorithm presented in the next section.

\section{Adaptations of MA-EDM to Improve its Long-Term Effectiveness}
\label{sec:adaptationofMAEDM}

As computing power increases and the instances grow, it is quite common for the effectiveness of algorithms to be affected~\cite{michel2005}.
Several options to deal with the increase in computing power, including the alternatives for expanding the number of generations or the size of the population have been analyzed~\cite{crestani2000soft}.
Based on these analyses, both theoretical and practical reasons are given in favor of increasing the size of the population and keeping the number of generations from growing excessively.
Alternatively, by introducing mechanisms that delay premature convergence, the number of useful generations can be increased~\cite{Crepinsek:13}.
For example, some optimizers that use Best Non-Penalized (BNP) strategy --- such as MA-EDM --- have been able to successfully leverage several thousand generations~\cite{Segura2016}.
However, even with this type of mechanism, there are limits on the number of useful generations, so with large computing capacities, some adaptations are still necessary. 

The options to take advantage of the increased computing power to manage larger populations or handle more generations do not entail any type of change in the algorithm.
A different option is to introduce computationally effective (and costly) strategies in some of the components that are used in each iteration of the algorithm.
In particular, for the improvement process, instead of applying a local search whose criterion is based on reaching a local optimum, a trajectory metaheuristic can be used to set a time-based stopping criterion.
This option is not promising at present for the LOP with large instances because with the current amount of power, it would require setting a very short time or evolving very few generations, but as the computing power increases, this option becomes viable.

Before being displaced by Memetic Algorithms, the Iterated Local Search (ILS) was one of the most successful strategies for the LOP~\cite{Schiavinotto:2004}, and to this day it continues to offer advantages in some variants of the LOP~\cite{Teran:11}, so we decided to resume this research and include it within MA-EDM.
Thus, we define the Memetic Algorithm with Explicit Management of Diversity and Enhanced Intensification (MA-EDM$ei$) with the novelty that in the intensification step, the local search is replaced with the ILS. 

\begin{algorithm}[!t]
\caption{Iterated Local Search (\textit{ILS})}
\
\label{alg:ILS}
\begin{algorithmic}[1]
\Require {\textit{ILS stopping criterion: $ils\_time$, \textit{Initial Solution:  $S_i$}}}
\State{\textbf{Intensification:} $S = Local\_Search(S_i)$}
\While {stopping criterion ($ils\_time$) not reached}
	\State{\textbf{(Perturbation)}}
	    \State{$S'$ = $S$}
        \For{i:\{1,2,3\}} (Perform 3 swaps)
            \State {$p_1$ = random position}
            \State {$p_2$ = random position}
            \State {swap($S'[p_1]$, $S'[p_2]$)}
        \EndFor
    \State{$S^{'}=Local\_Search(S^{'})$} \textbf{(Intensification)}
	\If{fitness($S^{'}$) $\ge$ fitness($S$)}
	   \State{$S=S^{'}$}
	\EndIf
\EndWhile
\State{\textbf{Return} $S$ (Best solution found)}
\end{algorithmic}
\end{algorithm}

In order to properly design an ILS strategy, we must select the perturbation and local search components.
Given the efficiency and effectiveness of the local search based on shifts, this was left unchanged.
In the case of the perturbation strategy, we analyzed several operators that have been successful in combinatorial optimization~\cite{Sabar:2015} and that exhibit different perturbation strengths.
In particular, Scramble, Inversion, Insertion and Swap were tested.
In the case of the swap strategy, we considered the version that incorporates the $p$ parameter to establish the number of 2-position swaps that are made~\cite{oliver1987study}.
The preliminary analyses allowed us to determine that the method based on performing multiple swaps was far superior.
In particular, $p = 3$ was enough to escape local optima, so the rest of the paper considers this operator exclusively.
In fact, this operator had already shown its superiority in~\cite{Teran:11}.

To extend MA-EDM (Algorithm~\ref{alg:MA_EDM}) and generate MA-EDM$ei$, lines 2 and 7 are updated by replacing the local search with ILS.
Algorithm~\ref{alg:ILS} details each step of ILS.
Each solution is intensified with local search after being perturbed, and it is accepted if it is at least as good as the best solution found.
In relation to the parallel version of this algorithm (PMA-EDM$ei$), it follows the same flow presented above (see Figure~\ref{fig:flow_chart_MAEDMILSP}).
The only difference is that the workers execute the ILS instead of the local search.

\section{Experimental Validation}
\label{sec:ExperimentalValidation}

The main objective of the experimental validation carried out in this research is to analyze the behavior of the leading LOP solvers in long-term executions, with the aim of understanding which design trends can help to better scale these approaches.
Additionally, the new solvers designed with the most promising design decisions are
applied to the most challenging LOP instances by simulating scenarios with a large amount of computation, in order to understand the room for improvement present compared to the current BKS~\cite{lugo2022diversity}, as well as to establish new levels that can be used in the future to validate new advances.
The experimental analysis is conducted from three perspectives. The first focuses on identifying the best approaches for extended long-term executions. The second examines the relationship between the components of MAs and their scalability. The final utilizes HPC to analyze the performance during extended long-term executions and to establish new BKS for future comparisons.

To account for the stochastic behavior of the algorithms analyzed, the runs were repeated 30 times with distinct random seeds, and the following statistical analyses were performed to compare the results~\cite{del2013srcs}.
First, the Kruskal Wallis test was applied as an omnibus test. 
If there were any significant differences, the Mann Whitney with Hommel's correction was used to identify the pair-wise differences. 
Additionally, with the aim of visualizing the results, the Nemenyi critical difference plot was used~\cite{Calvo:2016}. Significance levels were set to 0.05.
The experimental validation was carried out in a cluster whose nodes have two \textit{Intel Xeon E5 v4} and 32 GB of RAM.

\begin{table}[t]
\renewcommand{\arraystretch}{1.3}
\centering
\caption{Parameterization applied in each state-of-the-art solver}
\begin{tabular}{l l}
\hline
Optimizer & Parameterization \\ \hline
ILS$_r$ & $\epsilon=0.0001$, $i\_{change}=7$, $n\_{ni}=750$ \\
MA$_r$ & $N=200$ \\
CD-RVNS & Parameter-less Algorithm \\
MA-EDM & N=200, Crossover Operator=CX \\ \hline
\end{tabular}
\label{tab:parameterization}
\end{table}

\subsection{Relative performance between population-based algorithms and trajectory algorithms}

Recent best-known solutions for the LOP has been attained by considering population-based strategies.
In order to confirm the superiority of population-based approaches in long-term executions, this section analyzes the behavior of four algorithms: 
two trajectory-based algorithms (ILSr~\cite{Garcia:2019} and CD-RVNS~\cite{Santucci:2020}) and two population-based algorithms (MAr~\cite{Schiavinotto:2004} and MA-EDM~\cite{lugo2022diversity}). 
These runs were conducted by setting the stopping criterion to 192 hours (eight days).
Particularly, an instance of each size ($300$, $500$, $750$ and $1000$) of the xLOLIB2 benchmark set was selected randomly.
Table~\ref{tab:parameterization} shows the parameterization applied in each optimizer.

\begin{figure*}[!t] 
\centering 
\subfloat[N-stabu75\_750]{\includegraphics[width=0.45\textwidth]{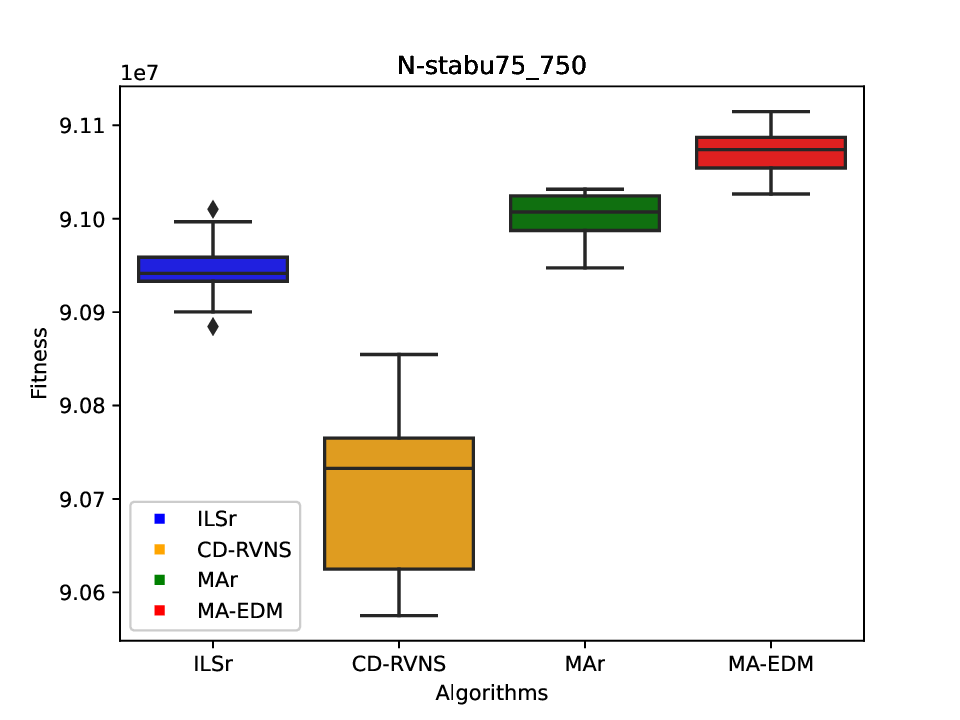}}
\subfloat[N-t70b11xx\_1000]{\includegraphics[width=0.45\textwidth]{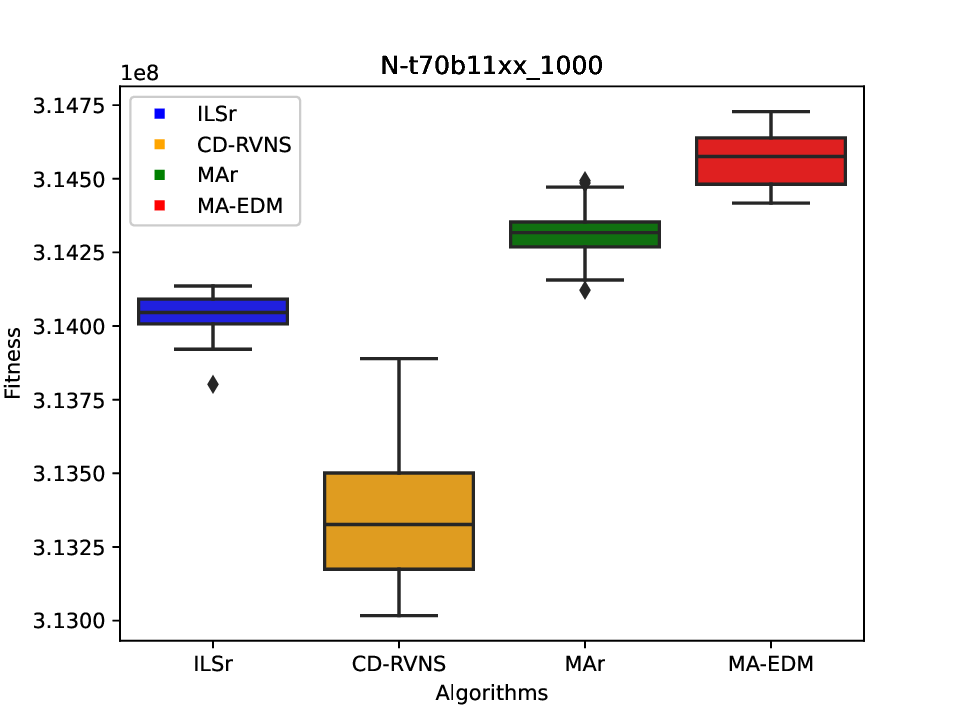}}
\qquad
\caption{Comparative analysis of trajectory-based and population-based algorithms with the stopping criterion set to 192 hours} 
\label{fig:TBvsPB}
\end{figure*}

The relative performance between the strategies was similar in all the instances.
Figure~\ref{fig:TBvsPB} shows the boxplots of the results for the two largest instances.
The results for the smallest instances can be found in the supplementary material.
%
%
%
MA-EDM consistently exhibits superior performance, outperforming both the other population-based algorithm and the trajectory-based algorithms and as the instances grow, the gap between the performance of MA-EDM and the rest of the strategies increase.
Note that the main difference between MAr and MA-EDM is the explicit management of diversity.
Thus, this experiment confirms the superiority of population-based strategies for long runs, as well as, the importance of correctly managing the diversity.

\subsection{Relationship between the scalability and components of MA-EDM}

In order to analyze the impact of increasing the number of generations, the performance of MA-EDM and MA-EDM$ei$ was analyzed using different stopping criteria (6, 24 and 192 hours).
Four instances of xLOLIB2 were chosen at random, one for each size available: 
300, 500, 750 and 1000.
The only parameter of MA-EDM is the population size ($N$).
Taking into account the results in~\cite{lugo2022diversity}, $N = 200$ was used.
In the case of MA-EDM$ei$, the same population size was used and the stopping criterion for ILS was set at 3.6 seconds. 
The preliminary analyses showed that after this run time, the improvements offered by ILS are quite limited.

\begin{figure*}[!t] 
\centering 
\subfloat[N-stabu75\_750]{\includegraphics[width=0.45\textwidth]{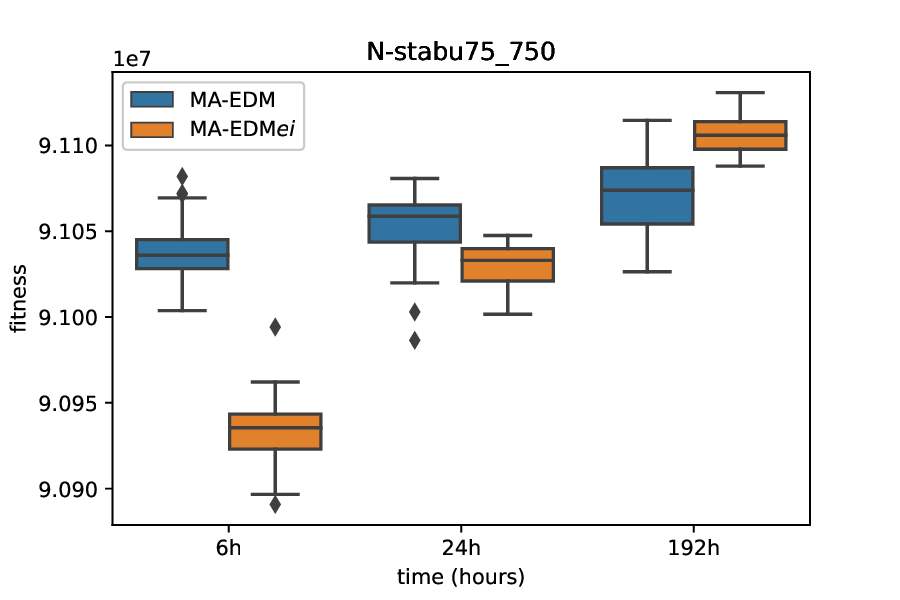}}
\subfloat[N-t70b11xx\_1000]{\includegraphics[width=0.45\textwidth]{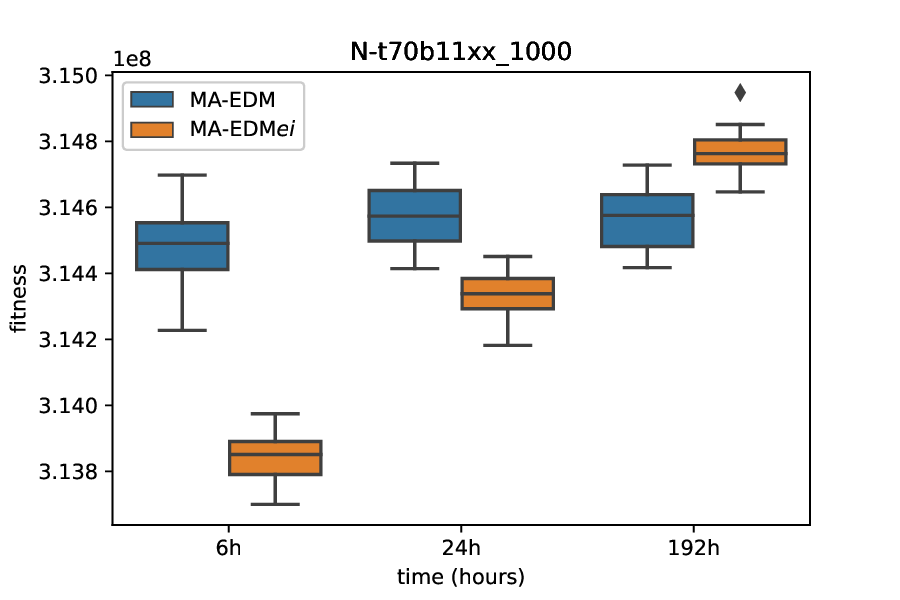}}
\qquad
\caption{Comparison between MA-EDM and MA-EDM$ei$ with three stopping criteria} 
\label{fig:maedmvsils}
\end{figure*}

Figure~\ref{fig:maedmvsils} shows the boxplots of the results obtained in the largest instances, while the results for the smallest ones can be consulted in the supplementary material.
With short stopping criteria, MA-EDM is superior.
This is especially clear in the 6-hour run, a case in which there are significant differences in favor of MA-EDM in every instance.
However, as the computing power increases, the improvements achieved by MA-EDM are limited, and it is surpassed by  MA-EDM$ei$ (also in a statistically significant way).
This improvement is most noticeable for larger instances. 
In the case of the instance of size $1000$, MA-EDM and MA-EDM$ei$ evolved on average $120711$ and $937$ generations, respectively.
The results show that as the computing capabilities increase, expanding the number of generations is not the best design choice.
In fact, in the case of the $1000$ instance, despite the noticeable margin for improvement still left for MA-EDM, 
there were no significant differences between the results obtained in 24 hours and 192 hours.

\begin{figure*}[!t] 
\centering 
\subfloat[N-stabu75\_750]{\includegraphics[width=0.45\textwidth]{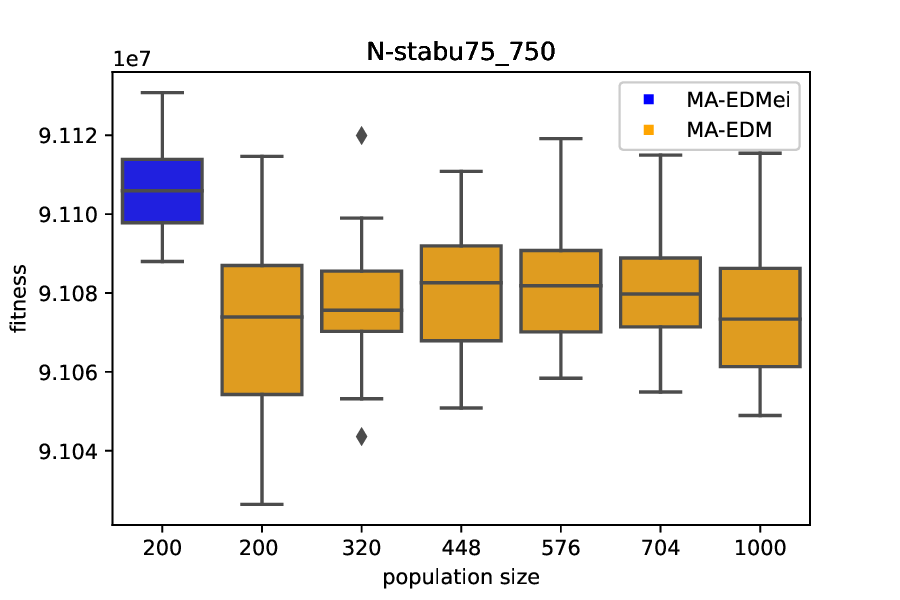}}
\subfloat[N-t70b11xx\_1000]{\includegraphics[width=0.45\textwidth]{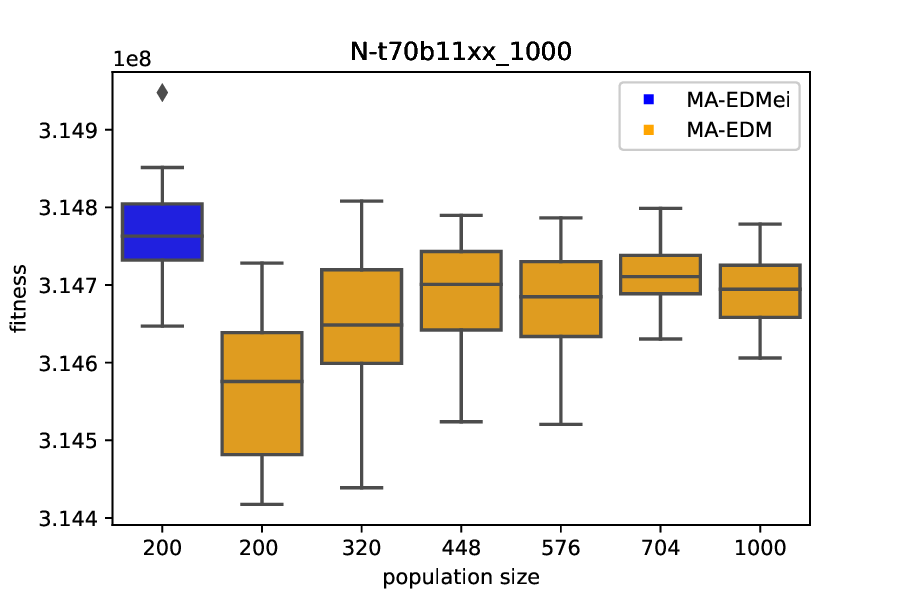}}
\qquad
\caption{Comparative analysis of population size in 192-hour runs for MA-EDM and MA-EDM$ei$}  
\label{fig:pop_size} 
\end{figure*}

Taking into account the analyses carried out in~\cite{crestani2000soft}, it is natural to consider MAs with large populations as an alternative to leverage the increase in computing power.
Because of this, MA-EDM was run by setting the stopping criterion to 192 hours and using larger population sizes.
In particular, 6 values were tested between $N = 200$ and $N = 1000$.
The boxplots of the results obtained for the instances N-stabu75\_750 and N-t70b11xx\_1000 are shown in Figure~\ref{fig:pop_size}.
The results obtained by MA-EDM$ei$ are also included.
The statistical tests carried out confirm that in both instances, there were significant differences between the results achieved by MA-EDM$ei$ and those obtained by MA-EDM, regardless of the population size used.
We conclude that in this case, decreasing the number of generations through population growth is not enough to take advantage of the computing resources; instead, more drastic changes must be made by modifying some of its internal components, as is done in MA-EDM$ei$.

\subsection{Analysis of performance in the very long-term}

\begin{figure*}[!t] 
\centering 
\subfloat[N-stabu75\_750]{\includegraphics[width=0.45\textwidth]{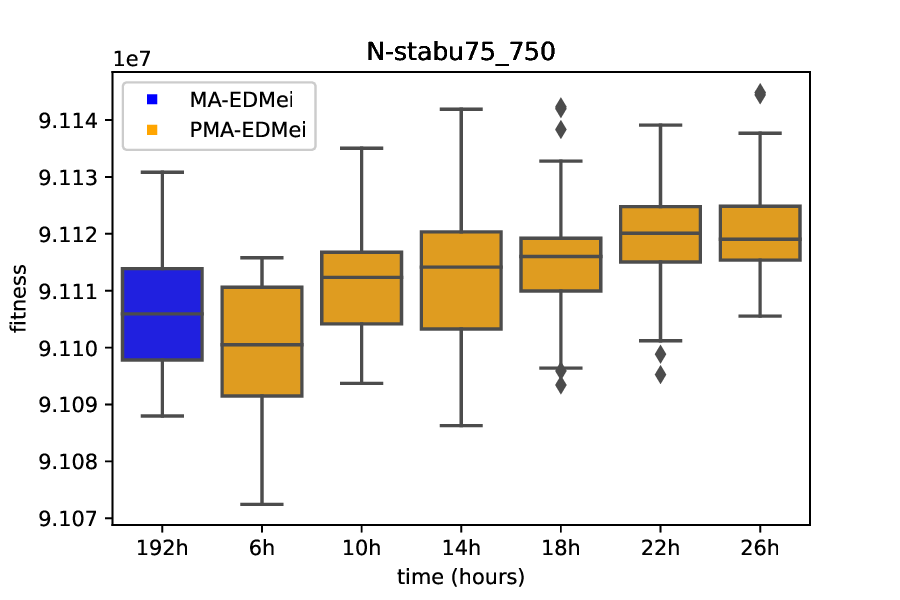}}
\subfloat[N-t70b11xx\_1000]{\includegraphics[width=0.45\textwidth]{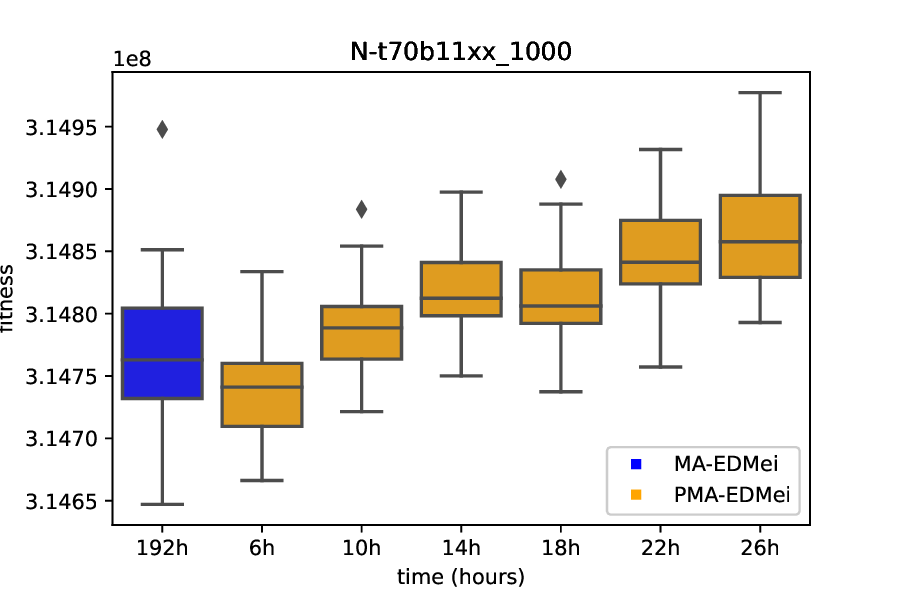}}
\qquad
\caption{Comparative performance analysis between PMA-EDM$ei$ and MA-EDM$ei$}  
\label{fig:pma-edmei_vs_ma-edmei} 
\end{figure*}

\begin{figure*}[!t]
\centering 
\subfloat[N-stabu75\_750]{\includegraphics[width=0.5\textwidth]{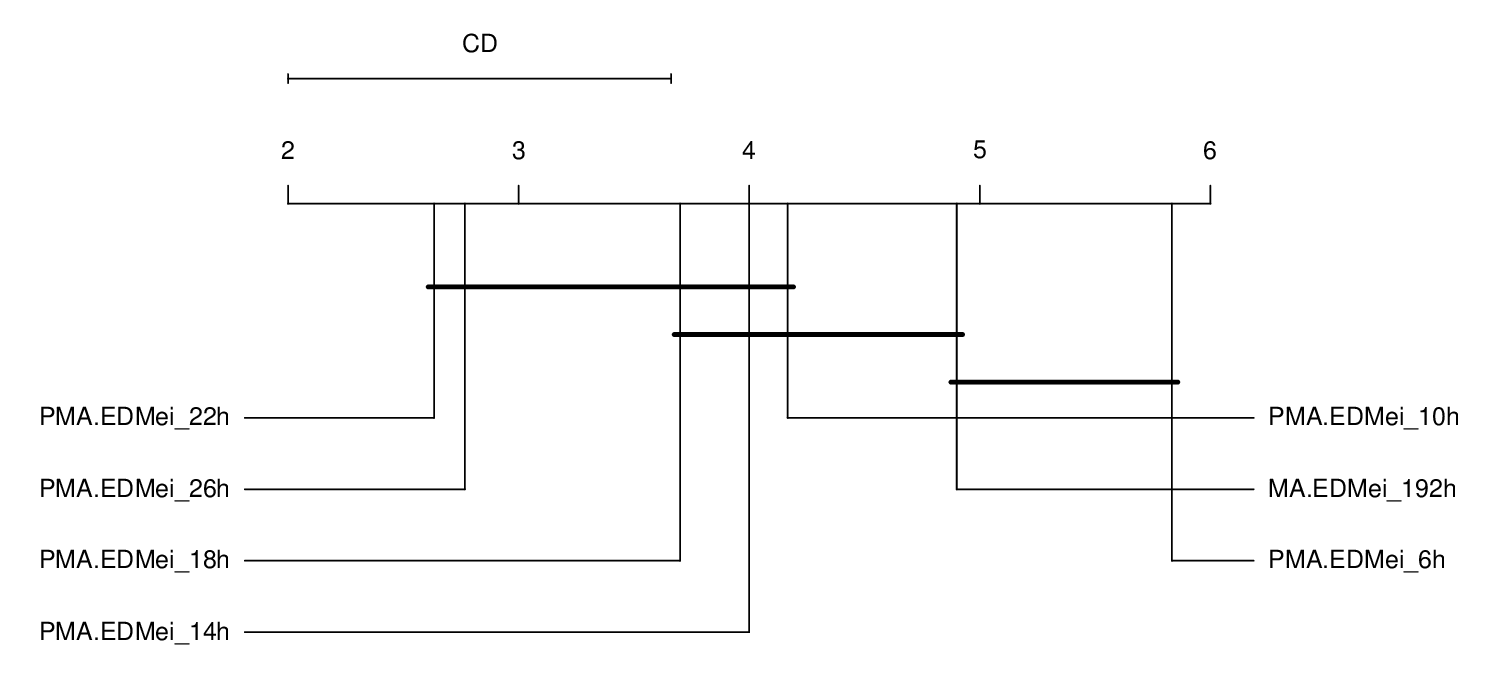}}
\subfloat[N-t70b11xx\_1000]{\includegraphics[width=0.5\textwidth]{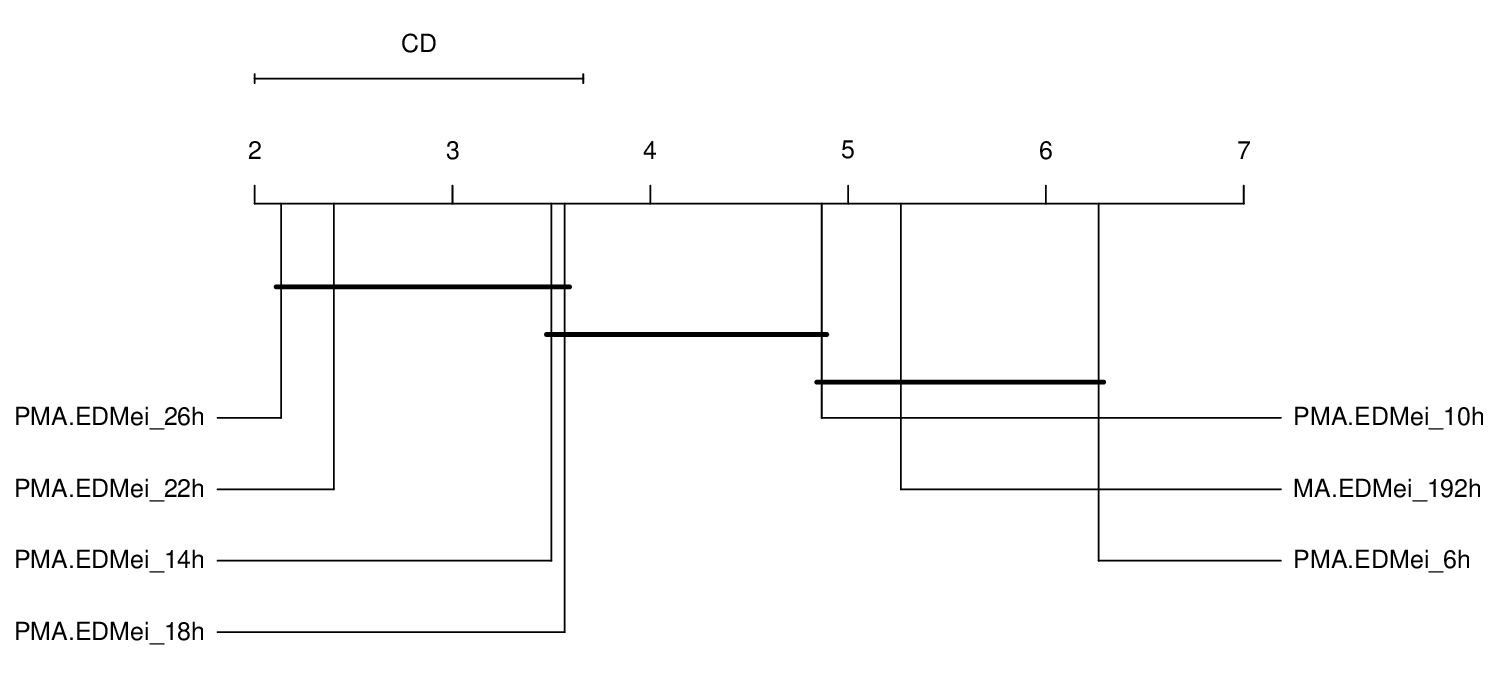}}
\qquad
\caption{Critical differences plot for MA-EDM$ei$ and PMA-EDM$ei$} 
\label{fig:critical_plots} 
\end{figure*}

This second experiment seeks to establish new BKS and analyze whether MA-EDM$ei$ scales appropriately when considering larger computation quantities than those used in the previous experiment.
Since the previous experiment involved optimizations lasting up to 8 days, it is not feasible to perform this new experiment using sequential executions.
Instead, PMA-EDM$ei$ with 32 worker processes was used.
One modification was made to the population size, consisting of establishing $N = 224$ in order for the population size to be a multiple of the number of workers.
Figure~\ref{fig:pma-edmei_vs_ma-edmei} shows the boxplots obtained by PMA-EDM$ei$ when executed setting the stopping time between 6 hours and 26 hours for the 750 and 1000 size instances that were used in the previous experiment.
Also shown are the results obtained by MA-EDM$ei$ at 192 hours.
The results for the 300 and 500 size instances are shown in the Supplementary Material.
Note that the computation hours used by PMA-EDM$ei$ range from $6 \times 32 = 192$ to $26 \times 32 = 832$ hours.
As expected, the 6-hour parallel results were worse than the 192-hour sequential results due to the loss of performance inherent in parallel models.
However, when using 10-hour runs, the median obtained by PMA-EDM$ei$ was higher than that of MA-EDM$ei$.
Therefore, the speedup is estimated to be between $\frac{192}{10} = 19.2$ and $\frac{192}{6} = 32$.
When considering the mean value, the efficiency is $\frac{25.6}{32} = 0.8$, and therefore each hour of parallel execution is estimated to correspond to $25.6$ hours of sequential execution.
Thus, this experiment simulates situations with current computing capabilities between 153 hours and 665 hours, i.e., practically one month.

In the case of the instance of size $1000$, the results continue to improve steadily, even on the longest runs.
The improvement in quality obtained compared to the previous BKS is immense, thus confirming that in instances of size $1000$ with the usual execution duration --- at most a few days --- there is considerable room for improvement.
Figure~\ref{fig:critical_plots} shows critical difference plots for these runs for the $750$ and $1000$ instances. These same figures for the case of 500 are shown in the Supplementary Material of this research. In the case of 300, it is not possible to construct this plot since the same result was obtained for all the executions carried out with both algorithms. 
In both instances, we see that the sequential run at 192 hours ranked between the parallel runs at 6 and 10 hours, confirming the previous findings in terms of the estimated efficiency.
In the 750 instance, 22-hour parallel runs were needed to significantly outperform the sequential results.
When a method generates near-optimal solutions, the resources allocated have to be greatly increased to achieve significant improvements, so this, coupled with the fact that the 22-hour run ranked similar to the 26-hour run, suggests that in this instance, very high quality solutions are being reached.
In the case of the $1000$ instance, the sequential results at 192 hours could be significantly outperformed by all the parallel runs starting at 14 hours.
In addition, the best average ranking was obtained at 26 hours, which is an indication that there is still room for improvement.

\begin{table*}[!t]
\renewcommand{\arraystretch}{1.3}
\centering
\caption{Results obtained with \textit{PMA-EDM$ei$} in 26-hour executions}
\begin{scriptsize}
\begin{tabular}{lrrrr}
\hline
Instance & \textit{Mean} & \textit{Best} & \textit{Worst} & Previous BKS \\
\hline
N-t65i11xx\_300    & 864234341.0    & \textbf{864234341}  & 864234341    & 864223106 \\
N-be75eec\_500     & 33489159.76    & \textbf{33489269}   & 33488116     & 33464804  \\
N-t59f11xx\_750    & 29202835.91    & \textbf{29208868}   & 29196802     & 29192928  \\
N-t59i11xx\_1000   & 3493537276.58  & \textbf{3493926328} & 3493030762   & 3491553089 \\
N-t65l11xx\_750    & 4948911.75     & \textbf{4949817}    & 4948170      & 4944869   \\
N-t65w11xx\_1000   & 72107414598.25 & \textbf{72127664540} & 72088824319  & 72045429648 \\
N-t75u11xx\_750    & 17369141424.0  & \textbf{17372345327} & 17364459304  & 17358027725 \\
N-t75u11xx\_1000   & 29732758325.58 & \textbf{29735696136} & 29727937321  & 29713770054 \\
N-tiw56n72\_750    & 77341956.33    & \textbf{77356607}   & 77324903     & 77300418   \\
N-tiw56r54\_1000   & 29320651.75    & \textbf{29324010}   & 29317329     & 29300654   \\
N-stabu75\_750     & 91121628.5     & \textbf{91144824}   & 91105532     & 91056055   \\
N-t70b11xx\_1000   & 314863785.0    & \textbf{314977289}  & 314792889    & 314603886  \\
N-tiw56r66\_750    & 35482250.91    & \textbf{35484408}   & 35477649     & 35461293   \\
N-tiw56r67\_1000   & 66711779.25    & \textbf{66724580}   & 66701153     & 66667742   \\
\hline
\end{tabular}
\end{scriptsize}
\label{tab:pmaedmei-moreinstances}
\end{table*}

In order to establish new levels that can be used in future research, PMA-EDM$ei$ was executed for a set of 10 additional instances, with the stopping criterion set at 26 hours.
Table~\ref{tab:pmaedmei-moreinstances} shows the mean, best and worst values achieved, as well as the BKS prior to this research~\cite{lugo2022diversity}.
This table also details the results obtained in the instances discussed previously.
Note that in every case, the previous BKS was surpassed.
In fact, even the worst result is better than the previous BKS.
Also worth noting is that in the $300$ and $500$ instances, the margin for improvement is relatively small, but grows to several million in the $1000$-size instances.

As a final experiment, and in order to set new BKS that are even more challenging to achieve, PMA-EDM$ei$ was executed with 6 instances and 32 worker processes, with the stopping criterion set at 120 hours.
These results are shown in Table~\ref{tab:exp3}.
In light of the ratios calculated earlier, these runs correspond to sequential optimizations lasting over 4 months.
Two of the instances are common to those used in the previous experiments, and the results continued to improve drastically in the case of $1000$, thus confirming that the current methods still have considerable room for improvement with instances of this size, and that in order to achieve solutions of this level of quality, truly long executions would be necessary.

\begin{table*}[!t]
\renewcommand{\arraystretch}{1.3}
\centering
\caption{Long-term experimental results (120 hours) obtained with \textit{PMA-EDM$ei$}}
\begin{scriptsize}
\begin{tabular}{lrrrr}
\hline
Instance            & \textit{Mean}    & \textit{Best}       & \textit{Worst}      & BKS         \\
\hline
N-stabu75\_750      & 91127572.35      & \textbf{91150565}   & 91105008            & 91056055    \\
N-t75e11xx\_750     & 823367015.6      & \textbf{823481759}  & 823055676           & 822729527   \\
N-usa79\_750        & 158342573.25     & \textbf{158388557}  & 158286792           & 158187746   \\
N-t70b11xx\_1000    & 314906869.75     & \textbf{314989031}  & 314817826           & 314603886   \\
N-t59n11xx\_1000    & 9624726.95       & \textbf{9626140}    & 9623842             & 9617008     \\
N-t70k11xx\_1000    & 28550270180.0    & \textbf{28558080100} & 28544793200         & 28520983800 \\
\hline
\end{tabular}
\end{scriptsize}
\label{tab:exp3}
\end{table*}


\section{Conclusion}
\label{sec:conclusion}

The design of metaheuristics has evolved over time due to the emergence of new ideas to perform optimization processes, as well as the continuous increase in computing capabilities.
This paper focuses on understanding the impact that the future increase in the availability of computational resources will have on Memetic Algorithms (MAs) in the particular case of the Linear Ordering Problem (LOP).
The use of parallelizations that keep the evolutionary model intact, as well as of long-term executions, made it possible to confirm that the MA that is used today to obtain the Best-Known Solutions (BKS) in the most challenging instances of the LOP (MA-EDM), as well as other state-of-the-art optimizers, are not able to adequately leverage the increase in computational resources on a large scale.
Two of the most common techniques used to take advantage of the increase in computational resources are to expand the size of the population or the number of generations.
However, with a high increase in computational resources, neither of these two techniques was successful in the case of the LOP.
A new heuristic optimizer (MA-EDM$ei$) was proposed based on integrating Iterated Local Search with MA-EDM, so as to avoid an excessive increase in the population size or in the number of generations, and thus increase the intensification potential.
This technique was able to successfully take advantage of amounts of computational resources similar to what a modern core can execute in more than 4 months, and yielded new solutions that significantly outperform the current BKS in the 18 instances that were used in this research.
These achievements confirm that as computational resources continue to increase, it will be necessary to adapt the current optimizers, and in particular it shows how promising it is to integrate MAs with techniques with great intensifying power.
Accordingly, we believe that in the future there will be a significant increase in MAs that will include intensification procedures with a high computational cost.
In fact, in recent years, the term matheuristics has become popular to refer to heuristics that include mathematical programming techniques in some of their components.
These types of techniques were not viable in the past due to their computational cost, but the number of areas in which this type of integration has been successful is growing.

In relation to the LOP, this research also shows that the current instances are still very challenging, and that there is still much room for improvement.
There is a need to integrate more methodological advances, mainly to be able to successfully deal with instances with matrices of size $500$ or greater, since in many fields it is not feasible to execute algorithms that are computationally equivalent to running a modern core for 4 months.
In addition, given that even with the longest executions carried out for this work, significant improvements continue to be obtained --- especially in the case of size $1000$ --- it is very likely that the optimal solutions can still be improved with respect to what is reported in this paper.

There are a large number of problems in which MAs very similar to those analyzed in this research make up the state-of-the-art methods.
In the future, it would be interesting to explore whether the same drawbacks in terms of scalability appear in these areas, and if the same types of modifications help alleviate these problems.
In relation to the LOP, it seems promising to further increase the capacities of the intensification process, so ideas related to the use of dynamic programming and branch-and-bound within the evolutionary process will be explored. 

\section*{Declarations}

\textbf{Confict of interest} The author does not have financial or non-financial interests that are directly or indirectly related to this article.

\bibliography{sn-bibliography}

\end{document}